\pdfoutput=1

\documentclass[11pt]{article}

\usepackage{EACL2023}

\usepackage{times}
\usepackage{latexsym}

\usepackage[T1]{fontenc}

\usepackage[utf8]{inputenc}

\usepackage{microtype}

\usepackage{inconsolata}
\usepackage{float}
\usepackage{hyperref}
\usepackage{url}
\usepackage{booktabs}       
\usepackage{amsfonts}       
\usepackage{nicefrac}       
\usepackage{microtype}      
\usepackage{xcolor}         
\usepackage{caption}
\usepackage{microtype}
\usepackage{graphicx}
\usepackage{subfigure}
\usepackage{multirow}
\usepackage{array}
\usepackage{makecell}
\usepackage{amsmath}
\usepackage{amssymb}
\usepackage{subfiles}
\usepackage{tabularx}
\usepackage[export]{adjustbox}%
\usepackage{wrapfig}
\usepackage{lipsum}
\usepackage{multicol}
\usepackage{pifont}
\usepackage{xspace}
\DeclareUnicodeCharacter{2212}{-}

\newcommand{\cmark}{\ding{51}}%
\newcommand{\xmark}{\ding{55}}%

\newcommand\bertbase{BERT$_{\normalsize {base}}$\xspace}
\newcommand\bertlarge{BERT$_{\normalsize {large}}$\xspace}

\title{Teacher Intervention: Improving Convergence of Quantization Aware Training for Ultra-Low Precision Transformers}

\author{Minsoo Kim\textsuperscript{1}, Kyuhong Shim\textsuperscript{2}, Seongmin Park\textsuperscript{1}, Wonyong Sung\textsuperscript{2} \and Jungwook Choi\textsuperscript{1}\thanks{\,\,\,Corresponding Author} \\
   \normalsize{\textsuperscript{1}Department of Electronic Engineering, Hanyang University} \\
   \normalsize{\textsuperscript{2}Department of Electrical and Computer Engineering, Seoul National University} \\
  \small{\texttt{\{minsoo2333, skstjdals, choij\}@hanyang.ac.kr}} \\
  \small{\texttt{\{skhu20, wysung\}@snu.ac.kr}} \\
  }

\begin{document}
\maketitle
\begin{abstract}
Pre-trained Transformer models such as BERT have shown great success in a wide range of applications, but at the cost of substantial increases in model complexity. Quantization-aware training (QAT) is a promising method to lower the implementation cost and energy consumption. However, aggressive quantization below 2-bit causes considerable accuracy degradation due to unstable convergence, especially when the downstream dataset is not abundant.
This work proposes a proactive knowledge distillation method called \textit{Teacher Intervention} (TI) for fast converging QAT of ultra-low precision pre-trained Transformers. TI intervenes layer-wise signal propagation with the intact signal from the teacher to remove the interference of propagated quantization errors, smoothing loss surface of QAT and expediting the convergence. Furthermore, we propose a \textit{gradual} intervention mechanism to stabilize the recovery of subsections of Transformer layers from quantization. The proposed schemes enable fast convergence of QAT and improve the model accuracy regardless of the diverse characteristics of downstream fine-tuning tasks. We demonstrate that TI consistently achieves superior accuracy with significantly lower fine-tuning iterations on well-known Transformers of natural language processing as well as computer vision compared to the state-of-the-art QAT methods. 
\end{abstract}

\section{Introduction}
\label{sec:introduction}

\begin{figure}[ht]
\begin{center}
\centerline{\includegraphics[width=0.85\columnwidth]{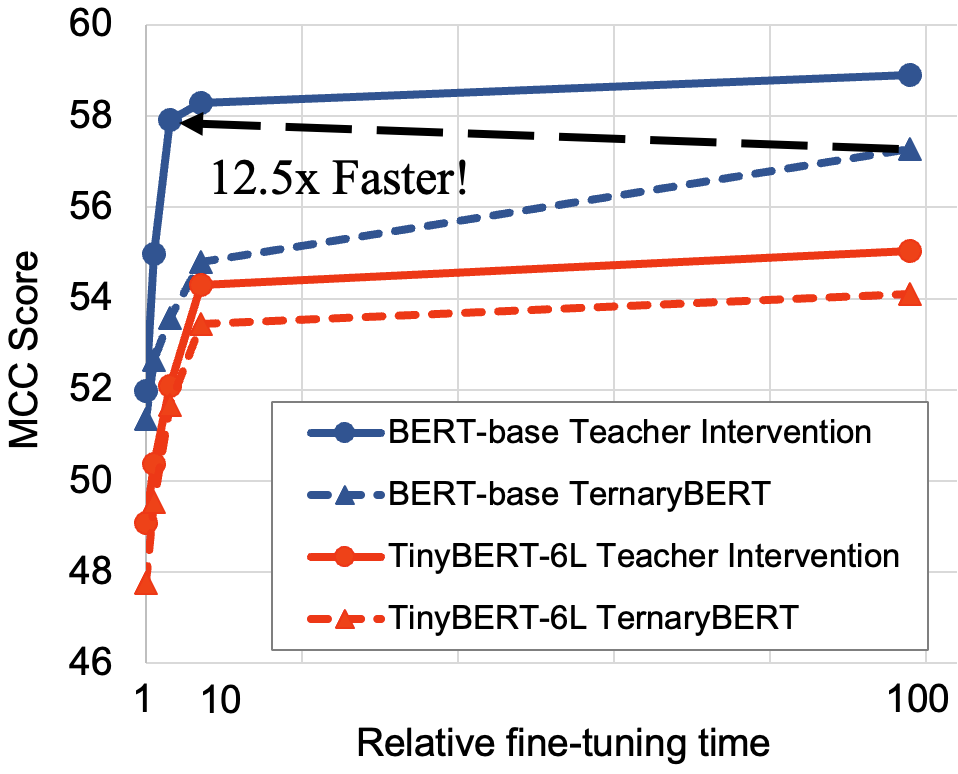}}
\caption{Comparison of fine-tuning time and accuracy of ternary weight quantized Transformer models between Teacher Intervention (TI) and TernaryBERT (BERT-base and TinyBERT-6L for CoLA task). TI achieves higher accuracy within 12.5$\times$ shorter fine-tuning time.}
\label{fig:training-time}
\end{center}
\vskip -0.4in
\end{figure}

The Transformer-based pre-trained neural networks have significantly improved the performance of various applications of artificial intelligence, including natural language processing (NLP) \citep{devlin2018bert, t5, gpt3} and computer vision (CV)~\citep{dosovitskiy2020image,touvron2021training,liu2021swin}. The self-attention mechanism represents these models \citep{vaswani2017attention}, which links different symbols within a sequence to obtain a relational representation. Thanks to the exceptional performance of the pre-trained Transformer models, there have been increasing needs for their efficient deployment. However, the gigantic size of the pre-trained Transformer models hinders straightforward implementation. 
Even relatively small models like BERT-base ~\citep{devlin2018bert} contain a few hundred million parameters,  
incurring profound memory and computation overhead for resource-constrained devices with limited memory and computing fabric. 
Therefore, seminal research efforts attempted to reduce this burden via model compression. \cite{behnke2020losing} and \cite{gordon2020compressing} pruned unimportant weights to reduce the number of parameters, while \cite{mao2020ladabert} further employed low-rank matrix factorization. 
In addition, Knowledge Distillation (KD)~\cite{kd} was employed in \citep{distilbert, pkd, mobilebert, minilm} to transfer knowledge of the original model (\textit{teacher}) to the compressed one (\textit{student}) by mimicking the Teacher's behavior.

Among many model compression techniques, \textit{quantization-aware training} (QAT) stands out for its recent success in reducing computational complexity and memory requirements of Transformer models \citep{bhandare2019efficient,q8bert,ibert}. QAT reflects quantization errors during the forward pass computation of stochastic gradient descent to train a more accurate quantized model. However, quantizing weight parameters of Transformers to a precision lower than 2-bits degrades the accuracy, especially when the dataset size for the target downstream tasks is not large enough \citep{ternarybert, binarybert}. 
Although few-sample fine-tuning of Transformer models has been reported to be highly unstable~\citep{lowft, ftweak, fewsampleft, fttechniques, ftstability}, efforts on understanding and improving QAT on small dataset tasks are limited. 
XTC~\citep{xtc_microsoft} recently revealed that a simple expansion in fine-tuning iterations could heal the QAT accuracy. Still, it caused an order-of-magnitude increase in QAT time, hindering the broad deployment of quantized Transformers.

This work proposes a proactive KD method called \textit{Teacher Intervention} (TI) for fast converging QAT of ultra-low precision pre-trained Transformers. We reveal that the difficulty of quantization on few-sample fine-tuning originates from disruption of loss surface due to quantization error propagation. To mitigate this undesirable phenomenon, we propose TI to intervene layer-wise signal propagation with the intact signal from the teacher. TI removes the interference of propagated quantization errors to smooth out the loss surface and expedite the convergence. We further discover that subsections of Transformer layers exhibit different susceptibility to quantization error for diverse downstream tasks. Thus, we propose a gradual intervention mechanism that first intervenes at the attention output for stable tuning of the feed-forward network, followed by self-attention map intervention for its recovery from quantization. The proposed gradual intervention along the subsections of Transformer layers enables fast convergence of QAT and improves the model accuracy regardless of the diverse characteristics of downstream fine-tuning tasks. 
We perform an extensive evaluation on various fine-tuned Transformers (BERT-base/large~\citep{devlin2018bert}, TinyBERT-4L/6L~\citep{tinybert}, and SkipBERT-6L~\citep{xtc_microsoft} for NLP, and ViT~\citep{dosovitskiy2020image} for CV) and demonstrate that TI consistently achieves superior accuracy with lower fine-tuning iterations compared to the state-of-the-art QAT methods. In particular, TI outperforms TernaryBERT~\citep{ternarybert} on GLUE tasks with 12.5$\times$ savings in fine-tuning hours, as shown in Fig.\ref{fig:training-time}.   
\section{Related Work}
\label{sec:related}

\subsection{Knowledge Distillation for BERT Compression}
\label{ssec:related_distillation}

Knowledge distillation (KD)~\cite{kd} is a transfer learning framework that passes on knowledge of a large model (\textit{teacher}) to a smaller one (\textit{student}).
Since KD provides extra guidance on how the student should behave, it can help mitigate accuracy degradation for model compression. Therefore, KD has been widely employed in training smaller BERT models for various application domains.

The most common distillation approach is to match the probability distribution from the final output softmax between the teacher and student for the same input, as shown in DistilBERT~\cite{distilbert}.
In addition to the distillation loss at the model output, PKD~\cite{pkd} suggested loss on intermediate output that matched the normalized output of the teacher and the student at each Transformer layer. MobileBERT~\cite{mobilebert} also employed per-head attention map transfer along with the customized network structure for constructing efficient Transformers. MiniLM~\cite{minilm} further transferred knowledge from the self-attention map as well as the value-relation. Considering the structural mismatch between the Teacher and Student models, MiniLM performed distillation only at the last Transformer layer. 

While the above mentioned studies focused on the task-agnostic BERT, there have been several efforts~\cite{tskd, irkd} to train tiny task-specific students. In this line of study, the task-specific, downstream fine-tuned BERT is first prepared, and the Student is trained with KD by utilizing this fine-tuned model as a teacher. As a hybrid approach, TinyBERT~\cite{tinybert} proposed a two-step KD, the first step for general distillation, followed by task-specific distillation. 

Although extensive research has been conducted to utilize KD for BERT compression, there are limited efforts in investigating and developing KD techniques for model quantization, an orthogonal model compression method. In this work, we develop a new KD technique primarily focusing on improving QAT of pre-trained Transformers. In particular, we reveal that more aggressive intervention of the teacher on the subsections of each Transformer layer helps the ultra-low precision model regain the model accuracy. 


\subsection{Quantization for Ultra-Low Precision BERT}

Quantization is a promising technique for reducing the high inference cost of large-scale models without changing the model structure.
Instead of representing numbers with the 32-bit floating-point (FP32) format, employing fixed-point representation, such as 8-bit integer (INT8) quantization, has achieved significant speedup and storage savings for BERT~\cite{q8bert, ibert}. However, direct quantization of weight parameters would suffer accuracy degradation of the original model accuracy when the quantization bit-precision is low. Therefore, quantization-aware training (QAT) is commonly applied for ultra-low precision model quantization. 

Recently, QAT has been applied for compressing BERT with precision lower than 2-bit. TernaryBERT~\cite{ternarybert} represents each weight element into one of three values $\{-1, 0, 1\}$. TernaryBERT actively incorporates KD into QAT for improving accuracy degradation. Especially, KD with the MSE loss on the attention score (before taking Softmax) and the output of each Transformer layer is employed for QAT. To further reduce the bit-precision, BinaryBERT~\cite{binarybert} suggested a modified QAT procedure that initializes the weights for binary quantization.
However, ternarizing or binarizing weight parameters significantly degrades the model accuracy, especially when the dataset size for the target downstream tasks is not large enough. 

In fact, it has been reported that finetuning BERT on downstream tasks with insufficient data is highly unstable~\cite{lowft, ftweak}. As a result, several works proposed modified finetuning procedures for improving the stability~\citep{fewsampleft, fttechniques, ftstability}. Still, the proposed approaches do not address the sensitivity of Transformer models on QAT for small datasets. 
XTC~\citep{xtc_microsoft} recently proposed a QAT method with significantly increased iterations and data augmentation to improve quantization accuracy of ultra-low bit precision Transformers. As illustrated in Fig.~\ref{fig:training-time}, however, this prolonged fine-tuning results in sizable deployment overhead, let alone costly data augmentation.
In this work, we discover that quantization significantly disrupts the propagation of self-attention in Transformer layers hindering the optimization process of QAT. Therefore, we propose a new KD-based method that proactively intervene the error propagation to improve convergence of QAT methods.


\section{Background and Motivation}
\label{sec:background}
\subsection{Transformer Layer}
\label{subsec:background_transformer}

The BERT model~\cite{devlin2018bert} is built with Transformer layers~\cite{vaswani2017attention}, which include two main sub-modules: Multi-Head Attention (MHA) and Feed-Forward Network (FFN). Input to the $l$-th Transformer layer is $\textbf{X}_l\in \mathbb{R}^{n\times d}$ where $n$ and $d$ are the sequence length and hidden state size, respectively. Let $H$ be the number of attention heads and $d_h=d/H$. $\textbf{W}^{Q}_h,\textbf{W}^{K}_h,\textbf{W}^{V}_h\in \mathbb{R}^{d\times d_h}$ are the weight parameters converting $\textbf{X}_l$ into Query ($\textbf{Q}=\textbf{X}_l \textbf{W}^{\textbf{Q}}_h$), Key ($\textbf{K}=\textbf{X}_l \textbf{W}^{\textbf{K}}_h$), and Value ($\textbf{V}=\textbf{X}_l \textbf{W}^{\textbf{V}}_h$), respectively. Then, attention score ($\textbf{AS}_h = QK^{\top}$), self-attention map ($\textbf{SA}_h = \text{Softmax}_h(\frac{\text{\textbf{AS}}_h}{\sqrt{d}}) $), and attention context ($\textbf{AC} = \text{\textbf{SA}}_h\textbf{V}$) can be defined.

The attention output (\textbf{AO}) is defined as $\textbf{AO}=\text{MHA}(\textbf{X}_l)=\text{Concat}(\text{\textbf{AC}}_1,\text{\textbf{AC}}_2,...\text{\textbf{AC}}_{N_H})\times \textbf{W}^O$.
Motivated by \cite{not_only_a_weight}, the attention output can be re-written per each token $i$:
\begin{equation}
\text{MHA}(\textbf{X}_l)(i) = \sum_{j=1}^{n} \alpha_{i,j}f(\textbf{X}_{l}(j)),
\label{eq:mha-decomp}
\end{equation}
where $f(x) := (x\textbf{W}^{V} + b^{V})\textbf{W}^{O}$ and $\alpha_{i,j}$ is $j$'th value of $i$'th token in SA$_h$. Therefore, the computation of MHA consists of two parts: self-attention generation (SA-GEN) corresponding to the attention map ($\alpha$) and self-attention propagation (SA-PROP) corresponding to $f(x)$. Fig.~\ref{fig:teacher-insertion} illustrates the structure of a Transformer layer indicating regions corresponding to SA-GEN and SA-PROP.
FFN consists of two fully-connected layers with weight parameters $\textbf{W}^1$ and $\textbf{W}^2$: 
\begin{equation}
\begin{aligned}
\text{FFN}(\textbf{Y}_l)=\text{GeLU}(\textbf{X}_l\textbf{W}^1+b^1)\textbf{W}^2+b^2.
\end{aligned}
\end{equation}
Therefore, an output of Transformer layer $X_{l+1}$ is defined as:
\begin{equation}
\begin{aligned}
\textbf{Y}_l=\text{LayerNorm}(\textbf{X}_l+\text{MHA}(\textbf{X}_l)), \\
\textbf{X}_{l+1}=\text{LayerNorm}(\textbf{Y}_l+\text{FFN}(\textbf{Y}_l)).
\end{aligned}
\end{equation}

\label{subsec:background_qat}

\subsection{Quantization-Aware Training}
Quantization-aware training (QAT) emulates inference-time quantization during fine-tuning of the full-precision model to adjust the parameters to be robust to the quantization error~\cite{sips_wonyong}~\cite{bnn_bengio}. In particular, ternary quantization represents all the weight parameters ($\textbf{W}^Q,\textbf{W}^K,\textbf{W}^V,\textbf{W}^O,\textbf{W}^1,\textbf{W}^2$) into ternary values $t\in\{+1,0,-1\}$ along with a scale factor $s$ for sub-2bit inference. In this work, we follow the approach of TWN~\citep{zhu2016trained} that analytically estimates the optimal scale factor $s$ and the ternary weight $\textbf{W}_t=\{t\}^{|\textbf{W}|}$ to minimize $\|\textbf{W}-s\cdot \textbf{W}_t\|$. 

Due to aggressive bit-reduction, ternary quantization causes significant accuracy loss. KD can help compensate for accuracy degradation, where the original full-precision model works as a teacher to guide the training of the quantized model as a student. In case of Transformer models, TernaryBERT~\cite{ternarybert} applied KD on every output activation $X_{l+1}$ as well as the attention score (\textbf{AS}) with mean squared error (MSE) loss: 
\begin{equation}
\begin{aligned}
L_{trm}=\sum_{l=1}^{L+1}\text{MSE}(\textbf{X}^S_l,\textbf{X}^T_l) + \sum_{l=1}^{L}\text{MSE}(\textbf{AS}^S_l,\textbf{AS}^T_l), 
\end{aligned}
\end{equation}
where $S$ and $T$ represent the student and teacher models, respectively.
Also, the output logits of the student ($P^S$) and the teacher ($P^T$) are used in TernaryBERT to compute the cross-entropy (CE) loss: 
\begin{equation}
\begin{aligned}
L_{pred}=\text{CE}(P^S,P^T).
\end{aligned}
\end{equation} 
We follow the settings of TernaryBERT as our baseline QAT method. 

\begin{figure}[ht]
\begin{center}
\centerline{\includegraphics[width=1\columnwidth]{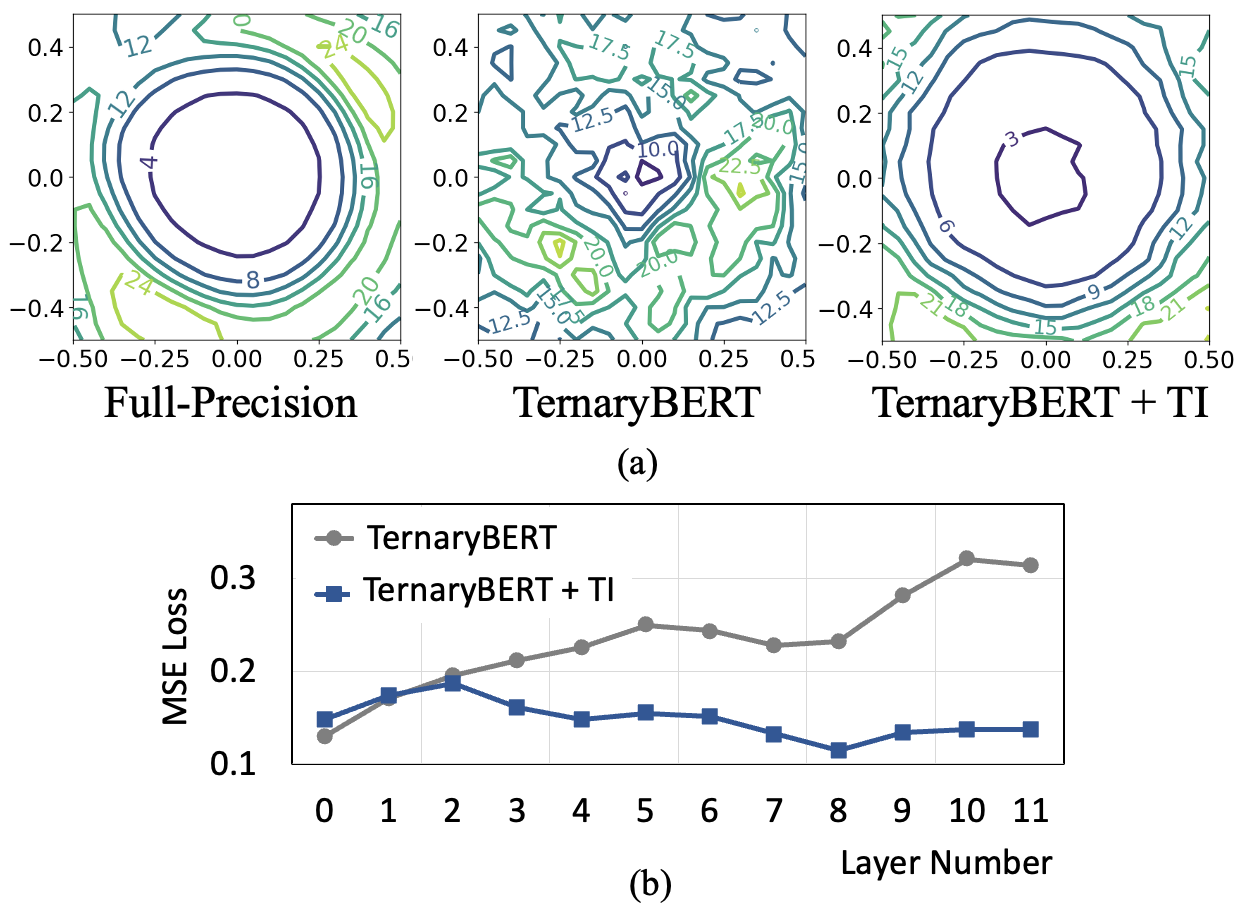}}
\caption{(a) Loss landscape visualization of a fine-tuned Transformer for a few-sample task (BERT-base, CoLA): A well-trained full-precision model and the quantized models without and with Teacher Intervention (TI). (b) MSE loss at the output of Transformer layers of TernaryBERT without and with TI}
\label{fig:loss_surface_layer_loss}
\vskip-0.4in
\end{center}
\end{figure}

\subsection{Challenges}
\label{subsec:background_challenge}
Despite attempts to bridge the accuracy gap, prior works on ultra-low precision Transformers~\citep{binarybert,ternarybert} still suffer noticeable accuracy degradation, especially when the dataset size is small. Recall \cite{ftstability} that the few-sample fine-tuning is unstable. But we observed that QAT often fails even if it fine-tunes from the successfully trained model. To gain intuition on this failure, we visualize the loss landscape of the quantized Transformers fine-tuned for a few-sample task (BERT-base, CoLA) in Fig.~\ref{fig:loss_surface_layer_loss}(a). In contrast to the smooth loss surface of the well-trained full-precision model, TernaryBERT exhibits sharp curvatures with many valleys, suffering unstable fine-tuning due to quantization. Investigating the internal behavior of Transformer layers under quantization elucidates the cause of unstable QAT. We measure the mean-square error (MSE) of the output of each Transformer layer between the full-precision baseline (teacher) and TernaryBERT (student). Fig.~\ref{fig:loss_surface_layer_loss}(b) reveals a prominent trend that MSE grows over the layers. (Similar trends can be observed in the other GLUE tasks.) This inflated error along the layers would degrade the model's accuracy. 

In this work, we focus on managing this aggravating impact of quantization errors on Transformers with a proactive knowledge distillation called \textit{Teacher Intervention} (TI). Interestingly, as shown in Fig.~\ref{fig:loss_surface_layer_loss}(b), TI successfully suppresses the error propagation and flattens the loss surface for favorable convergence without precipitously increased fine-tuning iterations. We discuss the technical details of TI in the next section.

\section{Method}
\label{sec:method}

\begin{figure*}[ht]
\begin{center}
\centerline{\includegraphics[width=2\columnwidth]{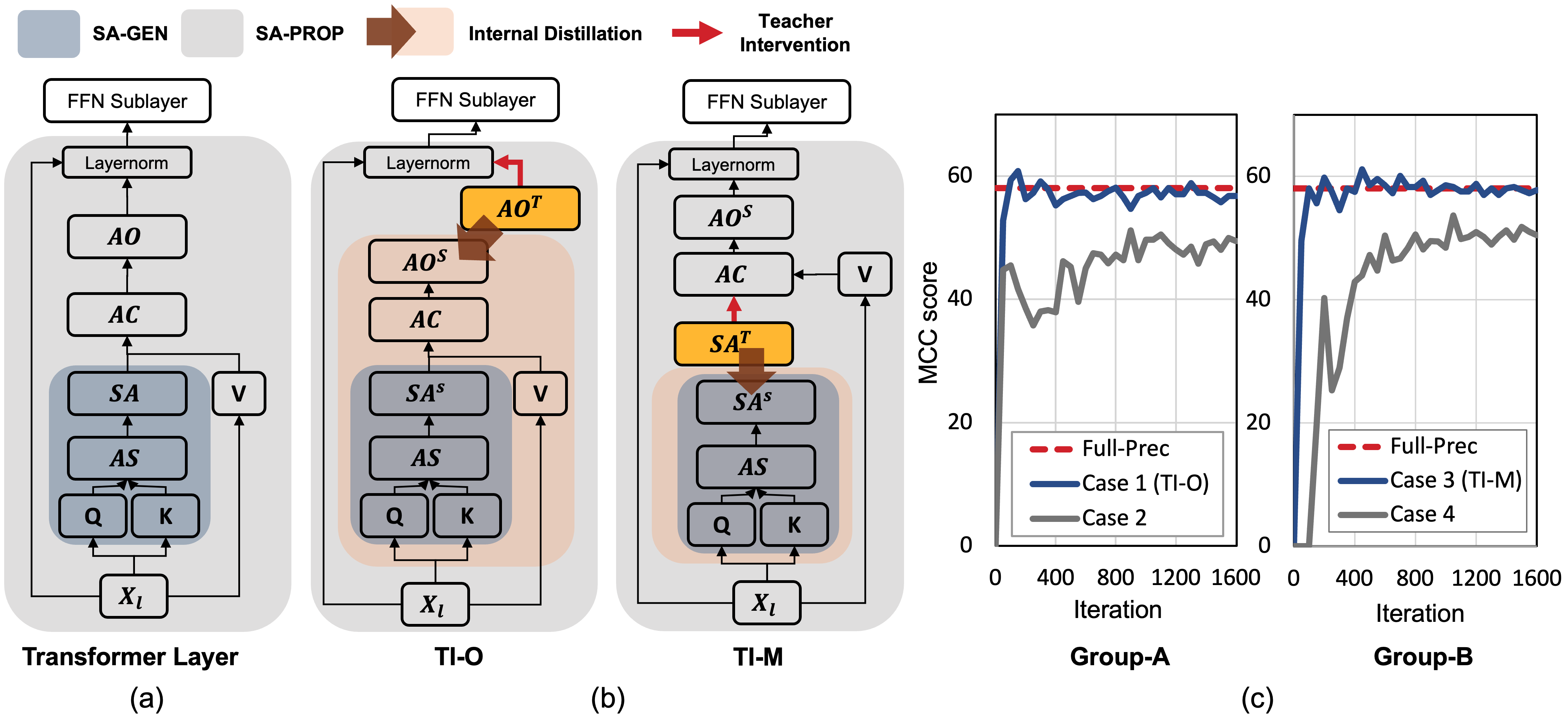}}
\caption{(a) Architecture of Transformer layer. (b) Locations of Teacher Intervention: output intervention (TI-O) and map intervention (TI-M). $\textbf{AO}^{T}$ and $\textbf{SA}^T$ are teacher's attention outputs. (c) Accuracy curves of controlled experiments. Group-A: Quantize all sub-layers with TI-O (Case 1) or do not quantize MHA (Case 2), Group-B: Quantize all sub-layers with TI-M (Case 3) or do not quantize SA-GEN (Case 4). Note rapid convergence for the cases with TI.}
\label{fig:teacher-insertion}
\end{center}
\vskip -0.35in
\end{figure*}

\subsection{Teacher Intervention}
\label{subsec:TI intro}

Teacher intervention (TI) is a KD method that aggressively intervenes in the student's signal propagation along the Transformer layers to suppress propagation of quantization error. Fig.~\ref{fig:teacher-insertion}(b) illustrates the two options for teacher intervention. First, intervention on the attention output (a.k.a. \textit{output intervention}, TI-O) replaces the student's attention output ($\textbf{AO}^S$ in Fig.~\ref{fig:teacher-insertion}(b)) in each Transformer layer with the teacher's ($\textbf{AO}^T$ in Fig.~\ref{fig:teacher-insertion}(b)). In this case, the FFN sub-layers are trained with ultra-low precision quantization without concerns of erroneous input from the preceding MHA. Meanwhile, the computation within the MHA sub-layer is quantized for internal distillation. Similarly, intervention on the self-attention map (a.k.a. \textit{map intervention}, TI-M) replaces the student's SA-GEN output ($\textbf{SA}^S$ in Fig.~\ref{fig:teacher-insertion}(b))with the teacher's ($\textbf{SA}^T$ in Fig.~\ref{fig:teacher-insertion}(b)). \footnote{We empirically investigated other options for teacher intervention and found that TI-O and TI-M were most representative.}

\begin{figure*}[ht]
\centering
\begin{center}
\centerline{\includegraphics[width=2\columnwidth]{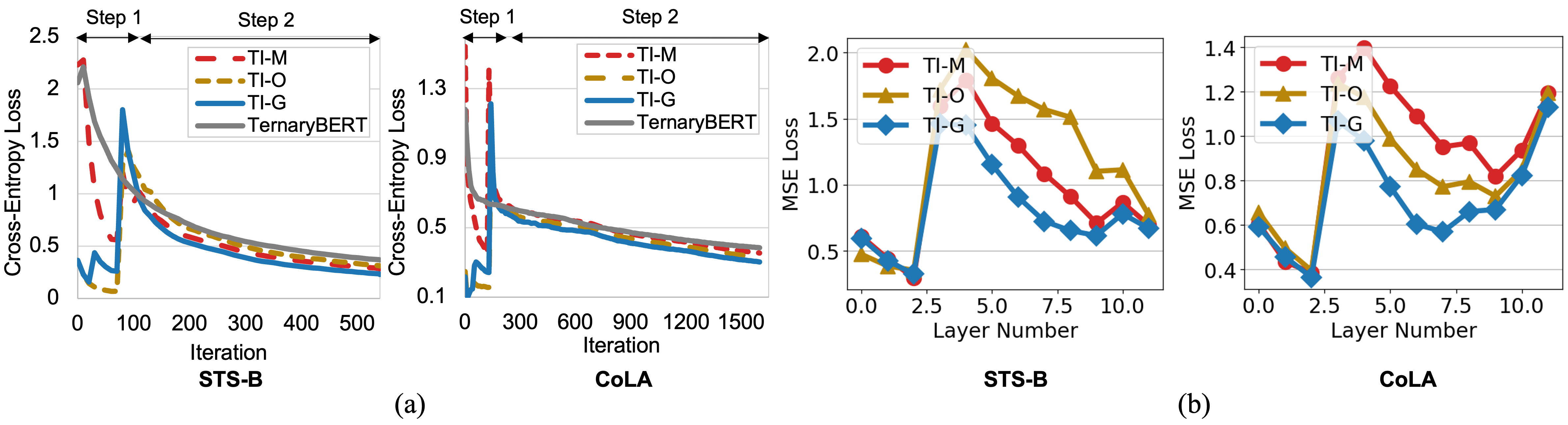}}
\vskip-0.1in
\caption{(a) Cross-entropy loss of QAT with different TI methods, and (b) Layer-wise MSE loss for different TI methods (Left: STS-B, Right: CoLA). Note that TI-G converges to lower loss than TI-M and TI-O on both tasks. }
\label{fig:transition_analysis}
\end{center}
\vskip-0.3in
\end{figure*}

The development of TI is motivated by the previous observation of the aggravating impact of quantization error along the layers (cf. Fig.~\ref{fig:loss_surface_layer_loss}(b)). We conjecture that the root cause of this phenomenon is error propagation instead of the quantization error itself. To confirm our hypothesis, we conducted controlled experiments for TI-O and TI-M with two groups of quantization cases: 

\begin{itemize}
    \item Group-A: Quantize all sub-layers with TI-O (Case1) or do not quantize MHA (Case2).
    \item Group-B: Quantize all sub-layers with TI-M (Case3) or do not quantize SA-GEN (Case4).
\end{itemize}

The key difference between the two groups is that Case2/4 propagates the quantization error through the sub-layers while Case1/3 does not, thanks to TI. Fig.~\ref{fig:teacher-insertion}(c) shows the convergence curves of the four cases on CoLA. As shown in the figure, Case1/3 converges rapidly to full-precision accuracy despite the ultra-low bit quantization in all the sub-layers. Whereas Case2/4 converges slowly to the sub-optimal point with noticeable accuracy degradation. Although Case2/4's MHA/SA-GEN computations are in full-precision, the error propagated from its preceding quantized FFN sub-layers still affects it, corrupting the attention output/self-attention map. On the other hand, TI-O and TI-M interrupt this error propagation to stabilize QAT on FFN sub-layers and MHA outputs, respectively. Therefore, Fig.~\ref{fig:loss_surface_layer_loss}(b) confirms that the error propagation disappears when TI is applied.

Motivated by this insightful observation, we devise a new QAT method that employs TI for the step-by-step reconstruction of sub-layers of Transformers. In the first step, TI is used to fine-tune the quantized weights of either FFN (TI-O) or SA-PROP (TI-M) sub-layers. Note that the convergence in this step is quick, as shown in Fig.~\ref{fig:teacher-insertion}(c). Therefore, Step1 takes only a fraction of fine-tuning epoch. In the second step, quantization is applied to the entire weights of Transformer layers for QAT. Since part of the parameters is already trained in Step1 with the guidance of TI, Step2 converges faster to a superior local minimum. In Sec.~\ref{sec:expr}, we demonstrate this improved convergence leads to boosts in accuracy for ultra-low bit Transformers.

\subsection{Gradual Teacher Intervention}
\label{subsec:TI-G}

\begin{table}[ht]
\setcellgapes{9pt}
\makegapedcells
\resizebox{1\linewidth}{!}{
\Huge
\begin{tabular}{cccccccccc}
\Xhline{5pt}
Type & \multicolumn{3}{|c|}{TI-M}     & \multicolumn{3}{c|}{TI-O}     & \multicolumn{3}{c}{TI-G} \\ \hline

Sub-layer & \multicolumn{1}{|c|}{GEN}  & \multicolumn{1}{c|}{PROP}     & \multicolumn{1}{c|}{FFN}    & \multicolumn{1}{c|}{GEN} & \multicolumn{1}{c|}{PROP}       & \multicolumn{1}{c|}{FFN}      & \multicolumn{1}{c|}{GEN}   & \multicolumn{1}{c|}{PROP} &
\multicolumn{1}{c}{FFN} \\ \hline
              
Step1-Phase1  &        \multicolumn{1}{|c|}{\multirow{2.5}{*}{\textbf{Q+TI}}}       & \multicolumn{1}{c|}{\multirow{2.5}{*}{Q}}       & \multicolumn{1}{c|}{\multirow{2.5}{*}{Q}}    & \multicolumn{2}{c|}{\multirow{2.5}{*}{\textbf{Q+TI}}}     & \multicolumn{1}{c|}{\multirow{2.5}{*}{Q}}    & \multicolumn{2}{c|}{\textbf{Q+TI}}                      & Q                  
\\ \cline{1-1} \cline{8-10} 

Step1-Phase2 & \multicolumn{1}{|c|}{}      & \multicolumn{1}{c|}{}     &  \multicolumn{1}{c|}{}     & \multicolumn{2}{c|}{}        &   \multicolumn{1}{c|}{}    &
\multicolumn{1}{c|}{\textbf{Q+TI}}      & \multicolumn{1}{c|}{Q}                  & \multicolumn{1}{c}{Q}                        \\ \hline

Step2        & \multicolumn{3}{|c|}{Q}     & \multicolumn{3}{c|}{Q}    & \multicolumn{3}{c}{Q}     \\ 
\Xhline{5pt}
\end{tabular}}
\caption{Quantization settings for teacher intervention. Different from TI-M and TI-O, gradual teacher intervention (TI-G) applies TI to quantization (Q) from large (GEN+PROP) to small (GEN) scope in Step1. \label{table:ti-settings}}
\vskip-0.1in
\end{table}

Given multiple teacher intervention options, which would achieve the best performance? In this section, we propose a unified approach that gradually applies the output intervention followed by the map intervention. Note that the two TI options have strengths and weaknesses. For example, TI-O focuses on tuning the FFN sub-layers, but it lacks consideration of the self-attention map recovery. On the other hand, TI-M is best suited for recovering the self-attention map, but it does not protect signal propagation through SA-PROP. Interestingly, we empirically discover that different downstream tasks of the pre-trained Transformers have diverse preferences; e.g., BERT-base fine-tuned on STS-B is sensitive to disruption in the self-attention map while the model fine-tuned on CoLA prefers careful tuning of the FFN sub-layers. Therefore, developing a unified solution that utilizes teacher intervention in various scopes is beneficial. As a natural combination, we propose a gradual teacher intervention mechanism that applies TI-O first to tune the FFN sub-layers (Step1-Phase1), followed by TI-M to recover the self-attention map (Step1-Phase2). The proposed method, called \textit{gradual intervention} (TI-G), has shown practical success in most ultra-low bit Transformers studied in this work. (We conducted an ablation study for the other possibilities of a unified solution for teacher intervention in ~\ref{subsec:ablation}.) Table.~\ref{table:ti-settings} summarizes the TI settings. Note that the two phases of TI-G in Step1 do not increase the total number of iterations thanks to fast convergence.

Fig.~\ref{fig:transition_analysis}(a) shows the convergence curves of different TI options for STS-B and CoLA. As discussed earlier, TI-M  shows better convergences than TI-O on STS-B, but the opposite trend is shown on CoLA. Nevertheless, TI-G always shows superior convergence compared to the other options, demonstrating its universal applicability. In particular, on STS-B, TI-G benefits from the output intervention in the first phase to favorably warm up the FFN sub-layers, and thus the map intervention in the next phase can reduce the loss more than TI-M. A similar situation happens in the case of CoLA. 

Fig.~\ref{fig:transition_analysis}(b) investigates the internal behavior of Transformers via the layer-wise MSE at the output of MHA. As discussed earlier, the fine-tuned Transformers exhibit distinct characteristics depending on the downstream tasks. For example, in the case of STS-B, TI-M is more effective in reducing MSE since the recovery of the self-attention map is essential. On the other hand, CoLA has preferred TI-O for its focus on tuning FFN sub-layers. In both cases, TI-G's gradual intervention with decreasing scopes from the attention output to the self-attention map helps achieve the smallest MSE. Therefore, this investigation suggests that the proposed unified intervention mechanism can manage diverse characteristics of fine-tuned Transformers for various downstream tasks. We demonstrate the general applicability of TI-G in the next section.

\section{Experiments}
\label{sec:expr}

In this section, we evaluate TI for QAT of fine-tuned BERT and vision Transformer with ternary weight quantization. We demonstrate that TI significantly boosts the convergence of QAT to achieve higher model accuracy within shorter fine-tuning time compared to the state-of-the-art QAT methods (TernaryBERT~\cite{ternarybert} and XTC~\cite{xtc_microsoft}) on various Transformer tasks and datasets. More extensive evaluation results on a wide range of BERT models as well as the ablation studies on various aspects of TI can be found in Appendix~\ref{subsec:ablation}.
Due to limited space, detailed experiment settings are also summarized in Appendix~\ref{subsec:experimental}.

\begin{table}[ht]
\centering
\resizebox{1\linewidth}{!}{
\begin{tabular}{lccccc}
        \toprule
        Iterations  & CoLA   & RTE       & MRPC    & STS-B  & Avg. Ratio    \\ \midrule
        Budget-O    & 1,603    & 233        & 343      & 538    & 1   \\
        Budget-A    & 13,325   & 4,468      & 7,050    & 10,066   & 16   \\
        Budget-C    & 159,909  & 53,621     & 84,611   & 120,795  & 200   \\        
        \bottomrule
    \end{tabular}}
\caption{Different budgets for fine-tuning iterations for GLUE tasks.\label{table:budgets}}
\vskip-0.2in
\end{table}

\begin{table*}[ht]
\centering
\resizebox{1\linewidth}{!}{
\begin{tabular}{llllllllll}
      &       & \multicolumn{4}{c}{\bertbase}     & \multicolumn{4}{c}{\bertlarge}    \\
             \cmidrule(lr){3-6} \cmidrule(lr){7-10}
     \multicolumn{1}{l}{QAT Type} & \multicolumn{1}{l}{Iterations} & \multicolumn{1}{l}{RTE} & \multicolumn{1}{l}{CoLA} & \multicolumn{1}{l}{STS-B} & \multicolumn{1}{l}{MRPC} &  \multicolumn{1}{l}{RTE} &
\multicolumn{1}{l}{CoLA} & \multicolumn{1}{l}{STS-B} & \multicolumn{1}{l}{MRPC}  \\
\toprule
Full-Prec & Budget-O        & 73.28        & 58.04                    & 89.24                     & 87.77                 & 70.39                   & 60.31                    & 89.83                     & 88.43                 \\ \midrule
 
TernaryBERT & Budget-O & 67.44 {\color{gray}\scriptsize±1.30}                  & 49.44    {\color{gray}\scriptsize±1.11}                  & 87.58    {\color{gray}\scriptsize±0.09}                   & 85.58 {\color{gray}\scriptsize±0.58}                   & 63.36    {\color{gray}\scriptsize±1.01}                 & 53.25       {\color{gray}\scriptsize±1.20}               & 88.65      {\color{gray}\scriptsize±0.16}                 & 88.31    {\color{gray}\scriptsize±0.20}                \\

TI-Map  & Budget-O   & 69.60       {\color{gray}\scriptsize±0.92}              & 51.37    {\color{gray}\scriptsize±1.23}                  & 87.75    {\color{gray}\scriptsize±0.12}                   & 86.25    {\color{gray}\scriptsize±1.03}                 & 66.13 {\color{gray}\scriptsize±1.12}                    & 52.40     {\color{gray}\scriptsize±1.65}                 & 88.61    {\color{gray}\scriptsize±0.16}                   & 88.67   {\color{gray}\scriptsize±0.32}                 \\

TI-Output  & Budget-O  & 69.31   {\color{gray}\scriptsize±0.57}                  & 50.91     {\color{gray}\scriptsize±0.94}                 & 87.76      {\color{gray}\scriptsize±0.22}                 & 86.04    {\color{gray}\scriptsize±0.61}             & 65.20             {\color{gray}\scriptsize±0.94}        & 52.66    {\color{gray}\scriptsize±1.27}                  & 88.56        {\color{gray}\scriptsize±0.16}               & 88.68    {\color{gray}\scriptsize±0.51}              \\

TI-G & Budget-O  & \textbf{70.32} {\color{gray}\scriptsize±0.72}           & \textbf{51.98} {\color{gray}\scriptsize±1.35}            & \textbf{87.77}   {\color{gray}\scriptsize±0.29}           & \textbf{86.44} {\color{gray}\scriptsize±0.49}     & \textbf{66.27}  {\color{gray}\scriptsize±0.79}          & \textbf{54.12}  {\color{gray}\scriptsize±1.13}           & \textbf{88.66}   {\color{gray}\scriptsize±0.05}           & \textbf{88.80} {\color{gray}\scriptsize±0.41}         \\ \midrule

TernaryBERT & Budget-O2 & 70.51 {\color{gray}\scriptsize±0.41}                  & 52.65    {\color{gray}\scriptsize±0.77}                  & 88.04   {\color{gray}\scriptsize±0.14}                   & 86.00 {\color{gray}\scriptsize±0.38}                   & 66.42    {\color{gray}\scriptsize±0.62}                 & 55.72 {\color{gray}\scriptsize±1.26}               & 89.00      {\color{gray}\scriptsize±0.09}                 & 88.22    {\color{gray}\scriptsize±0.82}                \\

TI-G   & Budget-O2 & \textbf{71.48} {\color{gray}\scriptsize±0.36}           & \textbf{54.98} {\color{gray}\scriptsize±0.66}            & \textbf{88.04}   {\color{gray}\scriptsize±0.18}           & \textbf{88.63} {\color{gray}\scriptsize±0.55}     & \textbf{68.11}  {\color{gray}\scriptsize±0.75}          & \textbf{57.55}  {\color{gray}\scriptsize±1.69}           & \textbf{89.12}   {\color{gray}\scriptsize±0.04}           & \textbf{88.25} {\color{gray}\scriptsize±0.46}         \\
\midrule
TernaryBERT  & Budget-O4 & 71.23 {\color{gray}\scriptsize±0.42}                  & 53.57    {\color{gray}\scriptsize±0.80}                  & 88.22    {\color{gray}\scriptsize±0.04}                   & 86.58 {\color{gray}\scriptsize±0.32}                   & 67.50    {\color{gray}\scriptsize±0.95}                 & 57.70       {\color{gray}\scriptsize±0.64}               & 89.12      {\color{gray}\scriptsize±0.01}                 & \textbf{89.12}    {\color{gray}\scriptsize±0.84}                \\

TI-G   & Budget-O4 & \textbf{73.16} {\color{gray}\scriptsize±0.36}           & \textbf{57.92} {\color{gray}\scriptsize±1.19}            & \textbf{88.48}   {\color{gray}\scriptsize±0.61}           & \textbf{89.56} {\color{gray}\scriptsize±0.52}     & \textbf{69.67}  {\color{gray}\scriptsize±0.72}          & \textbf{59.89}  {\color{gray}\scriptsize±1.07}           & \textbf{89.33}   {\color{gray}\scriptsize±0.10}           & 88.74 {\color{gray}\scriptsize±0.76}         \\
\midrule


  &  & \multicolumn{4}{c}{TinyBERT-4L}     & \multicolumn{4}{c}{TinyBERT-6L}    \\
             \cmidrule(lr){3-6} \cmidrule(lr){7-10}
     \multicolumn{1}{l}{QAT Type} & \multicolumn{1}{l}{Iterations} & \multicolumn{1}{l}{RTE} & \multicolumn{1}{l}{CoLA} & \multicolumn{1}{l}{STS-B} & \multicolumn{1}{l}{MRPC} & \multicolumn{1}{l}{RTE} &
\multicolumn{1}{l}{CoLA} & \multicolumn{1}{l}{STS-B} & \multicolumn{1}{l}{MRPC}\\
\toprule

Full-Prec & Budget-O        & 68.23                   & 43.06                    & 87.07                     & 87.76                                          & 74.00                   & 57.78                    & 88.74                     & 87.35                               \\ \midrule
TernaryBERT & Budget-O & 63.15  {\color{gray}\scriptsize±0.50}                 & 32.15 {\color{gray}\scriptsize±1.43}                   & 83.33    {\color{gray}\scriptsize±0.36}                 & 84.90  {\color{gray}\scriptsize±0.38}                                  & 68.74 {\color{gray}\scriptsize±1.42}                  & 47.77 {\color{gray}\scriptsize±0.35}                   & 87.29{\color{gray}\scriptsize±0.12}                     & 84.89 {\color{gray}\scriptsize±0.53}                         \\
TI-G & Budget-O  & \textbf{64.29}  {\color{gray}\scriptsize±0.72}        & \textbf{35.17} {\color{gray}\scriptsize±1.35}          & \textbf{83.58}   {\color{gray}\scriptsize±0.29}        & \textbf{85.48}   {\color{gray}\scriptsize±0.49}                      & \textbf{69.38}  {\color{gray}\scriptsize±0.78}        & \textbf{49.08}   {\color{gray}\scriptsize±1.24}        & \textbf{87.31}  {\color{gray}\scriptsize±0.11}          & \textbf{86.30}    {\color{gray}\scriptsize±0.63}    \\

\midrule

TernaryBERT & Budget-O2 & 64.74 {\color{gray}\scriptsize±0.76}                  & 34.49    {\color{gray}\scriptsize±1.89}                  & 84.10   {\color{gray}\scriptsize±0.34}                   & 85.73 {\color{gray}\scriptsize±0.05}                   & 69.67    {\color{gray}\scriptsize±0.95}                 & 49.54 {\color{gray}\scriptsize±0.22}               & 87.51      {\color{gray}\scriptsize±0.04}                 & 86.61    {\color{gray}\scriptsize±0.41}                \\

TI-G   & Budget-O2 & \textbf{64.98} {\color{gray}\scriptsize±1.45}           & \textbf{36.30} {\color{gray}\scriptsize±0.24}            & \textbf{84.27}   {\color{gray}\scriptsize±0.22}           & \textbf{86.44} {\color{gray}\scriptsize±0.31}     & \textbf{71.12}  {\color{gray}\scriptsize±0.63}          & \textbf{50.38}  {\color{gray}\scriptsize±0.56}           & \textbf{87.56}   {\color{gray}\scriptsize±0.09}           & \textbf{86.80} {\color{gray}\scriptsize±0.42}         \\
\midrule
TernaryBERT  & Budget-O4 & 65.10 {\color{gray}\scriptsize±0.55}                  & 35.91    {\color{gray}\scriptsize±0.31}                  & 84.36    {\color{gray}\scriptsize±0.26}                   & 86.70 {\color{gray}\scriptsize±0.13}                   & 70.75    {\color{gray}\scriptsize±0.36}                 & 51.66       {\color{gray}\scriptsize±0.51}               & 87.75      {\color{gray}\scriptsize±0.04}                 & 86.76    {\color{gray}\scriptsize±0.23}                \\

TI-G   & Budget-O4 & \textbf{65.22} {\color{gray}\scriptsize±0.55}           & \textbf{37.40} {\color{gray}\scriptsize±1.30}            & \textbf{84.39}   {\color{gray}\scriptsize±0.27}           & \textbf{87.22} {\color{gray}\scriptsize±0.13}     & \textbf{72.13}  {\color{gray}\scriptsize±0.12}          & \textbf{52.08}  {\color{gray}\scriptsize±0.01}           & \textbf{87.90}   {\color{gray}\scriptsize±0.14}           & \textbf{87.15} {\color{gray}\scriptsize±0.39}         \\
\bottomrule
\end{tabular}}
\caption{Accuracy comparison of QAT methods on BERT family (few-sample GLUE tasks) without and with TI for regular fine-tuning iterations (Budget-O/O2/O4). Each experiment is repeated 10 times.\label{tab:A-base-tiny}}
\vskip-0.1in
\end{table*}

\begin{table*}[ht]
\centering
\resizebox{1\linewidth}{!}{
\begin{tabular}{lllllllllll}

        &     & \multicolumn{4}{c}{TinyBERT-6L}   &  & \multicolumn{4}{c}{SkipBERT-6L}    \\
             \cmidrule(lr){3-6} \cmidrule(lr){8-11}
     \multicolumn{1}{l}{QAT Type} & \multicolumn{1}{l}{Iterations} & \multicolumn{1}{l}{RTE} & \multicolumn{1}{l}{CoLA} & \multicolumn{1}{l}{STS-B} & \multicolumn{1}{l}{MRPC} & \multicolumn{1}{l}{QAT Type} & \multicolumn{1}{l}{RTE} &
\multicolumn{1}{l}{CoLA} & \multicolumn{1}{l}{STS-B} & \multicolumn{1}{l}{MRPC}  \\
\toprule
Full-Prec   &  Budget-O     & 74.00        & 57.78                    & 88.74                     & 87.35                 &                    & 74.72                    & 55.37                     & 89.27           & 86.11                 \\
\midrule

TernaryBERT & Budget-A & 72.02 {\color{gray}\scriptsize±0.21}                  & 53.44    {\color{gray}\scriptsize±1.11}                  & 88.43    {\color{gray}\scriptsize±0.07}                   & 88.14 {\color{gray}\scriptsize±0.31}    & XTC               & 69.91   {\color{gray}\scriptsize±0.41}                 & 53.74       {\color{gray}\scriptsize±0.77}               & 88.77    {\color{gray}\scriptsize±0.03}                 & 86.29    {\color{gray}\scriptsize±0.57}                \\

TI-G   & Budget-A & \textbf{72.92} {\color{gray}\scriptsize±0.72}           & \textbf{54.29} {\color{gray}\scriptsize±1.35}            & \textbf{88.45}   {\color{gray}\scriptsize±0.29}           & \textbf{88.36} {\color{gray}\scriptsize±0.49}    & TI-G  & \textbf{70.87}  {\color{gray}\scriptsize±0.20}          & \textbf{56.46}  {\color{gray}\scriptsize±0.68}           & \textbf{88.94}   {\color{gray}\scriptsize±0.04}           & \textbf{86.98} {\color{gray}\scriptsize±0.44}         \\
\midrule
TernaryBERT  & Budget-C & 73.40 {\color{gray}\scriptsize±1.30}                  & 54.11    {\color{gray}\scriptsize±1.11}                  & \textbf{88.60}    {\color{gray}\scriptsize±0.02}                   & 88.43 {\color{gray}\scriptsize±0.58}           & XTC        & 73.76    {\color{gray}\scriptsize±0.54}                 & 56.30       {\color{gray}\scriptsize±0.67}               & 88.91      {\color{gray}\scriptsize±0.03}                 & \textbf{87.38 }  {\color{gray}\scriptsize±0.19}                \\

TI-G   & Budget-C & \textbf{73.82} {\color{gray}\scriptsize±0.41}           & \textbf{55.05} {\color{gray}\scriptsize±1.13}            & \textbf{88.60  } {\color{gray}\scriptsize±0.01}           & \textbf{88.62} {\color{gray}\scriptsize±0.02}    & TI-G & \textbf{74.48}  {\color{gray}\scriptsize±0.79}          & \textbf{56.32}  {\color{gray}\scriptsize±1.13}           & \textbf{88.92}   {\color{gray}\scriptsize±0.03}           & 87.34  {\color{gray}\scriptsize±0.41}         \\
\bottomrule

\end{tabular}}
\vskip-0.05in
\caption{Accuracy comparison of QAT methods on compressed BERT family (few-sample GLUE tasks) without and with TI for prolonged fine-tuning iterations (Budget-A/C). Each experiment is repeated 10 times. \label{tab:da-result}}
\vskip-0.15in
\end{table*}

\begin{table}[ht]
\resizebox{1\linewidth}{!}{
\begin{tabular}{lllll}
\toprule
    Iterations        & \multicolumn{2}{c}{Short (1K)}     & \multicolumn{2}{c}{Regular (10K/20K)}    \\
         \multicolumn{1}{l}{Dataset} & \multicolumn{1}{l}{CIFAR100} & \multicolumn{1}{l}{CIFAR10} & \multicolumn{1}{l}{CIFAR100} & \multicolumn{1}{l}{ImageNet}  \\ \midrule
         Full-Prec & \multicolumn{1}{l}{92.78} & \multicolumn{1}{l}{99.1} & \multicolumn{1}{l}{92.78} & \multicolumn{1}{l}{82.65} \\ \midrule
         TernaryBERT & 84.61 {\color{gray}\scriptsize±0.12}  & 97.32 {\color{gray}\scriptsize±0.02}  & 89.57 {\color{gray}\scriptsize±0.04}  & 75.40 {\color{gray}\scriptsize±0.12} \\
         
         TI-G & \textbf{85.28} {\color{gray}\scriptsize±0.04}  & \textbf{97.59} {\color{gray}\scriptsize±0.05}  & \textbf{90.07} {\color{gray}\scriptsize±0.04} &  \textbf{76.66} {\color{gray}\scriptsize±0.04}  \\
    
    \bottomrule
\end{tabular}}
\caption{Accuracy comparison of QAT methods on ViT-B for vision benchmarks.\label{tab:vit}}
\vskip-0.2in
\end{table}

\subsection{QAT Accuracy on Few-Sample BERT}
\label{subsec:bert}
First, we perform an extensive performance comparison of the proposed teacher intervention methods with the state-of-the-art QAT methods: TernaryBERT~\citep{ternarybert} and XTC~\citep{xtc_microsoft}. We evaluate TI on fine-tuned \bertbase (12 layer), \bertlarge (24 layer) ~\cite{devlin2018bert}, TinyBERT~\citep{tinybert} (6 layer), and SkipBERT ~\citep{xtc_microsoft} (6 layer) for GLUE tasks~\citep{glue}. We report the experimental results for the small dataset tasks (less than 10k of dataset size) here; the results on the other models and tasks are reported in Appendix~\ref{subsec:glue_large}.

To investigate the convergence of these QAT methods, we consider the following fine-tuning budgets:
\begin{itemize}
    \item Budget-O: The number of iterations reported in the original paper~\citep{ternarybert}. 
    \item Budget-O2: 2$\times$ iterations of Budget-O.
    \item Budget-O4: 4$\times$ iterations of Budget-O.
    \item Budget-A/C: prolonged fine-tuning budgets employed in XTC~\citep{xtc_microsoft}. 
\end{itemize}
A summary of fine-tuning budgets is shown in Table~\ref{table:budgets}. Note that the number of fine-tuning iterations of Budge-A is roughly 16$\times$ larger than Budget-O. In the following subsections, we categorize the experiments to two scenarios: few-sample fine-tuning (Scenario-1: Budget O/O2/O4) and prolonged fine-tuning (Scenario-2: Budget A/C).

\textbf{Scenario-1: Few-Sample Fine-tuning:}
Table~\ref{tab:A-base-tiny} summarizes the experimental results of few-sample fine-tuning. Consistent with the prior observations~\citep{xtc_microsoft}, the accuracy of TernaryBERT increased as the fine-tuning budget grew from Budget-O to Budget-O4. But all teacher intervention options have improved these baseline accuracies. Note that the preference between TI-M and TI-O varies across the tasks and models. For example, TI-M is mainly preferred on CoLA and RTE, while there is a marginally higher preference for TI-O on STS-B. For most cases, however, TI-G outperforms the other TI options. Although the accuracies of TI-M and TI-O are similar, TI-G's accuracy significantly exceeds both.
From these observations, we can conclude that the proposed gradual intervention significantly improves the QAT convergence for regaining model accuracy.

\textbf{Scenario-2: Prolonged Fine-tuning:}
Table~\ref{tab:da-result} summarizes the experimental results of prolonged fine-tuning. For fair comparisons, we followed the instructions of XTC to match the fine-tuning budgets, learning rates, and the model compression mechanism on both TinyBERT and SkipBERT. As expected, TI-G's accuracy is significantly increased from Budget-O to Budget-A/C than Budget-O4. Interestingly, TI-G's accuracy is higher than TernaryBERT and XTC for these prolonged fine-tuning budgets with noticeable margins. In fact, XTC's average accuracy on Budget-A (for SkipBERT-6L) is lower than TI-G's average accuracy on Budge-O4 (for TinyBERT-6L), highlighting the superior convergence of TI-G compared to XTC.

\begin{figure}[t]
\centering
{\includegraphics[width=1\columnwidth]{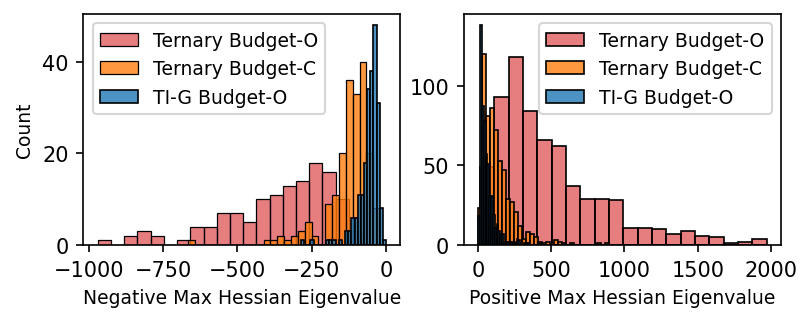}}
\caption{Convergence analysis of QAT methods with Hessian Max Eigenvalues.}
\label{fig:hessian}
\vskip-0.2in
\end{figure}

\subsection{QAT Accuracy on Vision Transformers}
\label{subsec:vit}
We further evaluate the proposed teacher intervention method on ViT. Table~\ref{tab:vit} summarizes the QAT accuracies of ViT fine-tuned for CIFAR10, CIFAR100, and ImageNet with the fine-tuning budgets following the original paper~\citep{dosovitskiy2020image}. As shown in the Table.~\ref{tab:vit}, TI-G outperforms TernaryBERT on all the QAT cases for ternary weight quantization. It is noteworthy that TI-G is exceptionally effective for fine-tuned ImageNet.

\subsection{Convergence Analysis}
\label{subsec:hessian_analysis}

In Sec.~\ref{sec:method}, we discussed that TI impedes propagation of quantization errors and flattens the loss surface for favorable convergence as shown in Fig.~\ref{fig:loss_surface_layer_loss}. To quantitatively analyze the convergence of QAT, we conducted the Hessian eigenvalue analysis ~\citep{park2021vision} on \,\bertbase\,  over CoLA task. As shown in Fig.~\ref{fig:hessian}, TernaryBERT suffers large magnitude (positive and negative) eigenvalues. The large magnitude Hessian eigenvalues indicate sharp loss surfaces, unfriendly for quantization~\cite{qbert,dong2019hawq}. A prolonged fine-tuning would suppress the large Hessian eigenvalues at the expense of significantly increased fine-tuning computations. In contrast, TI significantly reduces the magnitude with an order of magnitude smaller fine-tuning iteration. From this analysis, we can conclude that TI helps flatten the loss surface to improve the convergence of QAT.

\begin{figure*}[ht]
\begin{center}
\centerline{\includegraphics[width=2\columnwidth]{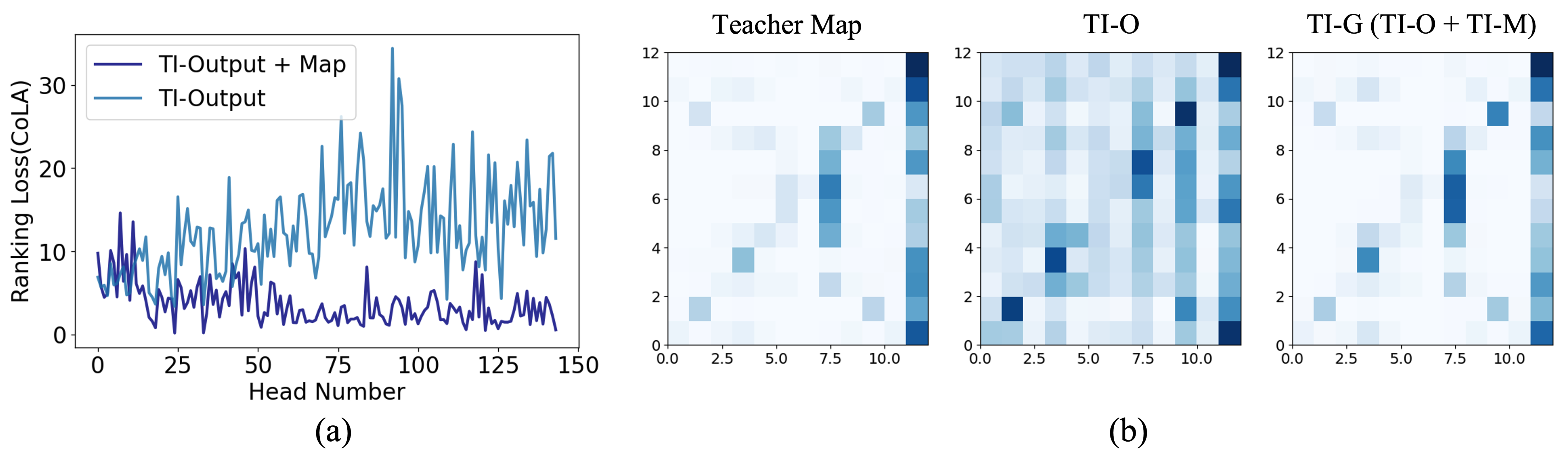}}
\caption{(a) Comparison of ranking loss between TI-O and TI-O with TI-M (TI-Output + Map) (b) Visualization of self-attention map with full-precision, TI-O and TI-G trained on CoLA. The input sentence is "John was lots more pleasure than Fred."\label{fig:self-attention-map}}
\end{center}
\vskip-0.2in
\end{figure*}

\subsection{Further Analysis of TI-G: Self-Attention Map Recovery}
\label{subsec:ti-m}

In this section, we provide an analysis of TI-M's effectiveness in the use of TI-G. Since the self-attention map encodes the relational representation of symbols within a sequence, quantization on SA-GEN would disrupt this relational encoding. Therefore, TI on the self-attention map (TI-M) is crucial for maintaining relative order within the self-attention map. As shown in Sec.~\ref{subsec:TI-G}, TI-O lacks consideration of the self-attention map recovery, requiring gradual teacher intervention mechanism that applies TI-M after TI-O for recovering the self-attention map. Fig.~\ref{fig:self-attention-map} shows two empirical evidences that TI-M helps to recover the self-attention map with TI-O. 

Fig.~\ref{fig:self-attention-map}(a) shows the ranking loss computed on the self-attention map of \bertbase after QAT on CoLA. Note that the ranking loss quantitatively indicates how much the relative order in the quantized self-attention map has been skewed from the teacher's self-attention map~\cite{ptq4vit}. As can be seen, QAT without TI-M leads to significant disruption of the self-attention map. In contrast, QAT with TI-M maintains the original ranking of cross-symbol correlation, successfully suppressing the ranking loss. We further visualize each quantized model's resulting self-attention maps in Fig.~\ref{fig:self-attention-map}(b). Contrary to using TI-O only, TI-O with TI-M successfully recovers the teacher self-attention map's distinctive feature. From this observation, we can conclude that TI-M approach used with TI-O is particularly beneficial for preserving the self-attention map under quantization.


\section{Conclusion}
\label{sec:conclusion}
In this work, we proposed a proactive knowledge distillation method for improving the convergence of QAT for ultra-low precision Transformers called teacher intervention. The proposed method intervenes in the propagation of quantization error to suppress accuracy degradation and improve QAT's convergence speed. We demonstrate that the proposed method outperforms the state-of-the-art QAT methods in achieving higher accuracy on various fine-tuned Transformers with smaller fine-tuning budgets. Our code is available at \href{https://github.com/MarsJacobs/ti-kd-qat}{https://github.com/MarsJacobs/ti-kd-qat}.

\section{Limitation}
\label{sec:limitation}
This work proposes a proactive knowledge distillation method called ~\textit{teacher intervention} for improving the convergence of QAT with small fine-tuning budgets. Although our analysis reveals that quantization error propagation is one of the main causes of sub-optimal convergence of QAT, a more in-depth investigation of the detailed recovery mechanism of QAT from quantization error would be interesting. Also, extending the discovery of proactive knowledge distillation to diverse Transformer architectures, including encoder-decoder and decoder-only models, would be a promising future research direction. 

\section*{Acknowledgements}
\label{sec:acknowledgement}
This work was partly supported by Samsung Advanced Institute of Technology, Samsung Electronics Co., Ltd, Institute of Information \& communications Technology Planning \& Evaluation(IITP) grant funded by the Korea government (MSIT) (2020-0-01297, Development of Ultra-Low Power Deep Learning Processor Technology using Advanced Data Reuse for Edge Applications) and National Research Foundation of Korea (NRF) grant funded by Korea government (MSIT) (No. 2021R1A2C1013513).

\bibliography{anthology,custom}
\bibliographystyle{acl_natbib}
\clearpage
\appendix

\section{Appendix}

In this section, we provide experimental setup and TI implementation in Sec.~\ref{subsec:experimental}.
We also provide evaluation results of GLUE
~\citep{glue} large datasets (SST-2, QNLI, MNLI, QQP) in Sec.~\ref{subsec:glue_large} and ablation studies in Sec.~\ref{subsec:ablation}. Further analysis of TI-G's effectiveness is provided in ~\ref{subsec:ti-m}.

\subsection{Experimental Settings}
\label{subsec:experimental}
\textbf{Models Description}
We use task-specific, fine-tuned BERT family models to evaluate our method. Further information on each model is as follows:

\begin{enumerate}
    \item \textbf{\bertbase} $L$=12, $d$=768 $N_H$=12, Contains about 110M parameters.
    \item \textbf{\bertlarge} $L$=24, $d$=1024 $N_H$=16, Contains about 340M parameters.
    \item \textbf{TinyBERT-4L} $L$=4, $d$=312 $N_H$=12, Contains about 14M parameters. This compressed model is fine-tuned using task-specific distillation method~\citep{tinybert} initialized from general pre-trained TinyBERT model file. \footnote{https://huggingface.co/huawei-noah/TinyBERT{\textunderscore}4L{\textunderscore}zh}
    \item \textbf{TinyBERT-6L} $L$=6, $d$=768 $N_H$=12, Contains about 67M parameters. This compressed model is fine-tuned using task-specific distillation method~\citep{tinybert} initialized from general pre-trained TinyBERT model file. \footnote{https://huggingface.co/huawei-noah/TinyBERT{\textunderscore}6L{\textunderscore}zh}
    \item \textbf{SkipBERT-6L} $L$=6, $d$=768 $N_H$=12, Contains about 67M parameters. This compressed model parameters are initialized from XTC ~\citep{xtc_microsoft} model initialization method: using every other layer of the fine-tuned {\bertbase}'s parameter to initialize the layer-reduced model.
\end{enumerate}

\noindent\textbf{Datasets and Settings}
We evaluate our method on GLUE benchmark ~\citep{glue}, which is a collection of resources for training, evaluating, and analyzing natural language understanding systems. For the training setting, we use the batch size of 16 for CoLA and 32 for other tasks. The learning rate starts from zero and gradually increases to 2e-5 during the warm-up stage and decays linearly to 2e-9 for total training epochs.

\begin{table}[H]
\centering
\resizebox{1\linewidth}{!}{
\begin{tabular}{lccc|cc}
\toprule
            &  \multicolumn{3}{c|}{~\cite{ternarybert}}    &     \multicolumn{2}{c}{~\cite{xtc_microsoft}} \\  \midrule
            &  Budget-O &  Budget-O2 &  Budget-O4 &  Budget-A &  Budget-C \\  \midrule
                
  DA       &   \xmark  &   \xmark  &   \xmark  &   \cmark  &   \cmark  \\
  Epoch & 3   &6 & 12 & 1 & 12 \\
\bottomrule
\end{tabular}}
\caption{Training budgets for the GLUE tasks. DA is data augmentation. \label{table:appen-budgets}}
\vskip-0.2in
\end{table}

\begin{table}[H]
\centering
\resizebox{1\linewidth}{!}{
\begin{tabular}{lcccc}
        \toprule
                         & CoLA   & RTE       & MRPC    & STS-B     \\ \midrule
        Train (noDA)     & 8,551   & 2,490      & 3,668    & 5,749      \\
        Train (DA)       & 213,212 & 142,991    & 225,630  & 322,121    \\ 
        DA / noDA        & 24.9   & 57.4      & 61.5    & 56.0    \\ \midrule
        Budget-O Time    & 259    & 97        & 72      & 223       \\
        Budget-A Time    & 2,149   & 1,862      & 1,484    & 4,160      \\
        Budget-C Time    & 25,792  & 22,342     & 17,813   & 49,915     \\
        \bottomrule
    \end{tabular}}
\caption{Training details for the GLUE tasks. DA is data augmentation. \label{table:appen-da}}
\end{table}

\noindent\textbf{Comparison of Training Cost}
We performed an extensive performance comparison in Sec.~\ref{subsec:bert} along the training budget to investigate the convergence of QAT methods. Table.~\ref{table:appen-budgets} shows the training setting of each budget with the Data Augmentation (DA) option, which is first proposed by ~\cite{ternarybert}~\cite{tinybert} artificially expanding the training sample for prolonged fine-tuning, causing a blow-up in fine-tuning costs. To investigate how much the DA magnifies the QAT overhead, we compare the number of training samples of GLUE small datasets and the number of data-augmented training samples in Table.~\ref{table:appen-da}. Furthermore, we measured the total QAT time for each Budget over GLUE small datasets with the A6000 single GPU, indicating that prolonged fine-tuning with DA would become profoundly painful, increasing model deployment costs and time.
\newline

\begin{table*}
\centering
\resizebox{1\linewidth}{!}{
\begin{tabular}{llllllllll}

        &     & \multicolumn{4}{c}{\bertbase}    & \multicolumn{4}{c}{\bertlarge}    \\
             \cmidrule(lr){3-6} \cmidrule(lr){7-10}
     \multicolumn{1}{l}{Method} & \multicolumn{1}{l}{Cost} & \multicolumn{1}{l}{SST-2} & \multicolumn{1}{l}{QNLI} & \multicolumn{1}{l}{MNLI} & \multicolumn{1}{l}{QQP}  & \multicolumn{1}{l}{SST-2} &
\multicolumn{1}{l}{QNLI} & \multicolumn{1}{l}{MNLI} & \multicolumn{1}{l}{QQP}  \\
\toprule
   &  Full-Prec     & 93.57        & 91.32                    & 84.59                     & 89.34                 & 93.57                   & 92.29                    & 86.49                     & 89.55                            \\
\midrule

TernaryBERT & Budget-O & 91.82 {\color{gray}\scriptsize±0.21}                  & 90.68    {\color{gray}\scriptsize±1.11}                  & 83.74    {\color{gray}\scriptsize±0.07}                   & 89.09 {\color{gray}\scriptsize±0.31}             & 91.62   {\color{gray}\scriptsize±0.41}                 & 91.91       {\color{gray}\scriptsize±0.77}               & 85.52    {\color{gray}\scriptsize±0.03}                 & 89.26   {\color{gray}\scriptsize±0.57}                \\

TI-Gradual   & Budget-O & \textbf{92.16} {\color{gray}\scriptsize±0.72}           & \textbf{90.91} {\color{gray}\scriptsize±1.35}            & \textbf{84.13}   {\color{gray}\scriptsize±0.29}           & \textbf{89.11} {\color{gray}\scriptsize±0.49}    & \textbf{92.12}  {\color{gray}\scriptsize±0.20}          & \textbf{92.11}  {\color{gray}\scriptsize±0.68}           & \textbf{86.11}   {\color{gray}\scriptsize±0.04}           & \textbf{89.46} {\color{gray}\scriptsize±0.44}         \\
\midrule

        &     & \multicolumn{4}{c}{TinyBERT-4L}    & \multicolumn{4}{c}{TinyBERT-6L}    \\
             \cmidrule(lr){3-6} \cmidrule(lr){7-10}
     \multicolumn{1}{l}{Method} & \multicolumn{1}{l}{Cost} & \multicolumn{1}{l}{SST-2} & \multicolumn{1}{l}{QNLI} & \multicolumn{1}{l}{MNLI} & \multicolumn{1}{l}{QQP}  & \multicolumn{1}{l}{SST-2} &
\multicolumn{1}{l}{QNLI} & \multicolumn{1}{l}{MNLI} & \multicolumn{1}{l}{QQP}  \\
\toprule
   &  Full-Prec     & 92.86        & 91.32                    & 84.59                     & 89.34                 & 93.57                   & 92.29                    & 86.49                     & 89.55                            \\
\midrule

TernaryBERT   & Budget-O & 91.43 {\color{gray}\scriptsize±0.72}           & 84.18 {\color{gray}\scriptsize±1.35}            & 80.93   {\color{gray}\scriptsize±0.29}           & 85.02 {\color{gray}\scriptsize±0.49}    & 91.12  {\color{gray}\scriptsize±0.20}          & 89.71  {\color{gray}\scriptsize±0.68}           & 83.51 {\color{gray}\scriptsize±0.04}           & 89.43 {\color{gray}\scriptsize±0.44}         \\

TI-Gradual   & Budget-O & \textbf{91.45} {\color{gray}\scriptsize±0.72}           & \textbf{84.51} {\color{gray}\scriptsize±1.35}            & \textbf{81.06}   {\color{gray}\scriptsize±0.29}           & \textbf{85.09} {\color{gray}\scriptsize±0.49}    & \textbf{91.43}  {\color{gray}\scriptsize±0.20}          & \textbf{89.81}  {\color{gray}\scriptsize±0.68}           & \textbf{83.64}   {\color{gray}\scriptsize±0.04}           & \textbf{89.44} {\color{gray}\scriptsize±0.44}         \\
\bottomrule

\end{tabular}}
\caption{Accuracy comparison of QAT methods on BERT family (large-sample GLUE tasks) without and with TI for
regular fine-tuning iterations (Budget-O). Each experiment is repeated 3 times. \label{bert_family_glue_large}}
\end{table*}

\begin{table*}
\centering
\resizebox{1\linewidth}{!}{
\begin{tabular}{llllllllll}

        &     & \multicolumn{4}{c}{\bertbase}    & \multicolumn{4}{c}{\bertlarge}    \\
             \cmidrule(lr){3-6} \cmidrule(lr){7-10}
     \multicolumn{1}{l}{Method} & \multicolumn{1}{l}{Cost} & \multicolumn{1}{l}{RTE} & \multicolumn{1}{l}{CoLA} & \multicolumn{1}{l}{STS-B} & \multicolumn{1}{l}{MRPC} &  \multicolumn{1}{l}{RTE} &
\multicolumn{1}{l}{CoLA} & \multicolumn{1}{l}{STS-B} & \multicolumn{1}{l}{MRPC}  \\
\toprule
   & Full-Prec        & 73.28        & 58.04                    & 89.24                     & 87.77                 & 70.39                   & 60.31                    & 89.83                     & 88.43                 \\ \midrule

TernaryBERT & Budget-A & 73.16 {\color{gray}\scriptsize±0.54}                  & 54.81    {\color{gray}\scriptsize±0.39}                  & 88.92    {\color{gray}\scriptsize±0.02}                   & \textbf{87.77} {\color{gray}\scriptsize±0.32}                   & 70.39   {\color{gray}\scriptsize±0.21}                 & 58.65       {\color{gray}\scriptsize±0.31}               & 89.77 {\color{gray}\scriptsize±0.01}                 & \textbf{88.91}    {\color{gray}\scriptsize±0.59}                \\

TI-G   & Budget-A & \textbf{74.24} {\color{gray}\scriptsize±1.36}           & \textbf{58.29} {\color{gray}\scriptsize±0.50}            & \textbf{89.01}   {\color{gray}\scriptsize±0.01}           & 87.74 {\color{gray}\scriptsize±0.96}      & \textbf{71.48}  {\color{gray}\scriptsize±0.95}          & \textbf{59.24}  {\color{gray}\scriptsize±1.38}           & \textbf{89.89}   {\color{gray}\scriptsize±0.07}           & 88.52 {\color{gray}\scriptsize±0.22}         \\
\midrule
TernaryBERT  & Budget-C & 73.40 {\color{gray}\scriptsize±0.41}                  & 58.25    {\color{gray}\scriptsize±0.84}                  & 89.23 {\color{gray}\scriptsize±0.01}                   & 88.43 {\color{gray}\scriptsize±0.31}                   & 71.48    {\color{gray}\scriptsize±0.06}                 & 59.11       {\color{gray}\scriptsize±1.47}               & 89.85      {\color{gray}\scriptsize±0.01}                 & \textbf{89.08}   {\color{gray}\scriptsize±0.55}                \\

TI-G   & Budget-C & \textbf{75.93} {\color{gray}\scriptsize±0.90}           & \textbf{58.91} {\color{gray}\scriptsize±0.54}            & \textbf{89.39} {\color{gray}\scriptsize±0.16}           & \textbf{87.85} {\color{gray}\scriptsize±0.26}     & \textbf{72.68}  {\color{gray}\scriptsize±1.50}          & \textbf{59.17}  {\color{gray}\scriptsize±0.61}           & \textbf{90.10}   {\color{gray}\scriptsize±0.01}           & 88.74  {\color{gray}\scriptsize±0.41}         \\
\bottomrule

\end{tabular}}
\caption{Accuracy comparison of QAT methods on BERT family (few-sample GLUE tasks) without and with TI for prolonged fine-tuning iterations (Budget-A/C). Each experiment is repeated 10 times. \label{table:bert_base_large_da}}
\end{table*}

\noindent\textbf{Teacher Intervention Implementation}
Our experiments were performed on A6000 GPUs. We use Pytorch 1.10.2 for the implementation of TI and this implementation is based on the TernaryBERT Pytorch code base. \footnote{https://github.com/huawei-noah/Pretrained-Language-Model/tree/master/TernaryBERT} We would like to emphasize that TI has strong advantages of simplicity for implementation. To implement TI operation, we modified the teacher model forward operation, enforcing the teacher model to return layer-wise MHA outputs and SA-GEN outputs which will be used in TI-O and TI-M respectively. When the teacher model's layer-wise required outputs are returned, we passed these to the student model forward operation's input. With these additional inputs, we replace the student's attention map/output with the teacher's in the attention sub-layer forward operation. More detailed implementation can be found at \href{https://github.com/MarsJacobs/ti-kd-qat}{https://github.com/MarsJacobs/ti-kd-qat}.

\begin{table}
\centering
\label{table:albation_decompose}
\begin{adjustbox}{width=1\linewidth}
\begin{tabular}{lcccc}
\toprule
   &  \bertbase &  \bertlarge &  TinyBERT-4L &  TinyBERT-6L \\ \midrule 
Baseline &  72.51 & 73.29 & 65.88 & 72.17 \\ \midrule
TI - output loss &  73.92 & 73.79 & 66.85 & 72.77 \\
TI - two step &  73.95 & 74.26 & 66.32 & 72.59 \\
TI &  \textbf{74.13} &\textbf{ 74.46} & \textbf{67.13} & \textbf{73.02} \\
\bottomrule
\end{tabular}
\end{adjustbox}
\caption{Ablation study of Teacher Intervention. Each result is averaged score over GLUE small datasets (RTE, CoLA, STS-B, MRPC)}
\label{table:ablation_ti}
\vskip -0.2in
\end{table}

\subsection{Additional Experimental Results}
\label{subsec:glue_large}
\textbf{Large-Sample Fine-tuning}
Table.~\ref{bert_family_glue_large} shows evaluation of TI in GLUE large datasets (SST-2, QNLI, MNLI, QQP). TI-G outperforms the baseline for all the cases as we discussed in Table.~\ref{tab:A-base-tiny}, showing that TI improve convergence of QAT in large sample fine-tuning.
\newline

\noindent\textbf{Additional Results of Prolonged Fine-Tuning}
Table.~\ref{table:bert_base_large_da} shows additional evaluation results of TI in prolonged fine-tuning (\bertbase and \bertlarge with Budget-A/C). As can be seen, TI-G shows superior accuracy to TernaryBERT with a noticeable margin in Budget-A/C both. Furthermore, TI-G achieves close to full-precision accuracy with Budget-A, showing TI's beneficial effect on the convergence of QAT.

\begin{table}
\centering
\label{table:ti_methods}
\begin{adjustbox}{width=1\linewidth}
\begin{tabular}{lcccc}
\toprule
            &  \bertbase &  \bertlarge &  TinyBERT-4L &  TinyBERT-6L \\ \midrule 
Inverted TI & 74.27 & 74.37 & 68.28 & 72.98 \\
Stochastic TI & 74.32 & \textbf{74.57} & 68.17 & 72.93 \\
TI-G & \textbf{74.37} & 74.41 & \textbf{68.32} &\textbf{ 73.06} \\
\bottomrule
\end{tabular}
\end{adjustbox}
\caption{Comparison of unified TI strategies. Each result is averaged score over GLUE small datasets (RTE, CoLA, STS-B, MRPC)}
\label{table:ablation_ti-G}
\vskip-0.2in
\end{table}

\subsection{Ablation Study}
\label{subsec:ablation}

\noindent\textbf{TI two step QAT and output loss}
In this section, we delve deeper into evaluating the efficacy of the TI method. As TI-O employs one additive loss for internal distillation (Attention Output MSE loss, called output loss) and TI QAT is performed in two steps, conducting an ablation study on TI (exploiting output loss and two-step QAT separately) provides a deeper understanding of our method. Table.~\ref{table:ablation_ti} demonstrates the effect of output loss and the use of a two-step method with TI independently with BERT family models over GLUE small datasets. As can be seen, both factors act positively on the performance of quantized models. Interestingly, two-step QAT factor provides a powerful impact (Row 2) in compressed models (TinyBERT-4L and TinyBERT-6L), and output loss shows a distinct impact (Row 3) in the \bertlarge model. When both factors are implied together (Row 4), performance is boosted further.  \\
\newline
\noindent\textbf{Exploration of unified teacher intervention methods} \, In this section, we investigate the best unified teacher intervention strategy. We proposed a unified intervention approach (TI-G) in Sec.~\ref{subsec:TI-G}, applying TI with output intervention (TI-O) followed by map intervention (TI-M). As more scheduling choices are left when using TI-O and TI-M together, we propose additional two scheduling methods, Inverted TI, and Stochastic TI. Inverted TI applies TI-O and TI-M in the opposite order of TI-G, and stochastic TI means choosing TI-O and TI-M options randomly in every training iteration, ensembling each TI option's effect in a stochastic way. Table.~\ref{table:ablation_ti-G} shows the comparison between unified TI scheduling methods with BERT family models over GLUE small datasets. TI-G outperforms other unified TI scheduling methods in almost every model, which shows our proposed teacher intervention unification approach's efficacy.
\label{sec:appendix}

\end{document}